\documentclass[11pt,letterpaper]{article}
\usepackage{naaclhlt2015}
\usepackage{times}
\usepackage{latexsym}
\usepackage{mathtools}
\usepackage{url}
\usepackage{amsmath}
\usepackage{multirow}
\usepackage[normalem]{ulem}
\usepackage{color}
\usepackage[usenames,dvipsnames]{xcolor}
\usepackage{enumitem}
\usepackage[lined,boxed,commentsnumbered]{algorithm2e}
\usepackage{graphicx}
\usepackage{epstopdf}
\usepackage{subfig}
\usepackage{mdwlist}
\usepackage{enumitem}
\usepackage{amsfonts}
\usepackage{amsmath}
\usepackage{mathtools}
\usepackage{setspace}

\title{Socially-Informed Timeline Generation for Complex Events}

\author{Lu Wang ~~~~  Claire Cardie ~~~~ Galen Marchetti\\
  Department of Computer Science \\
  Cornell University \\
  Ithaca, NY 14853 \\
  {\tt \{luwang, cardie\}@cs.cornell.edu} ~~~~ {\tt gjm97@cornell.edu} \\}

\begin{document}
\maketitle

\begin{abstract}
\fontsize{10}{12}\selectfont
Existing timeline generation systems for complex events consider only information from traditional media, ignoring the rich social context provided by user-generated content that reveals representative public interests or insightful opinions.
We instead aim to generate socially-informed timelines that contain both news article summaries and selected user comments. 
%
%
We present an optimization framework designed to balance topical cohesion between the article and comment summaries along with their informativeness and coverage of the event. 
%
Automatic evaluations on real-world datasets that cover four complex events show that our system produces more informative timelines than state-of-the-art systems. In human evaluation, the associated comment summaries are furthermore rated more insightful than editor's picks and comments ranked highly by users. 

\end{abstract}

\section{Introduction}
\begin{figure}[t]
\hspace{-3mm}
\includegraphics[width=87mm,height=54mm]{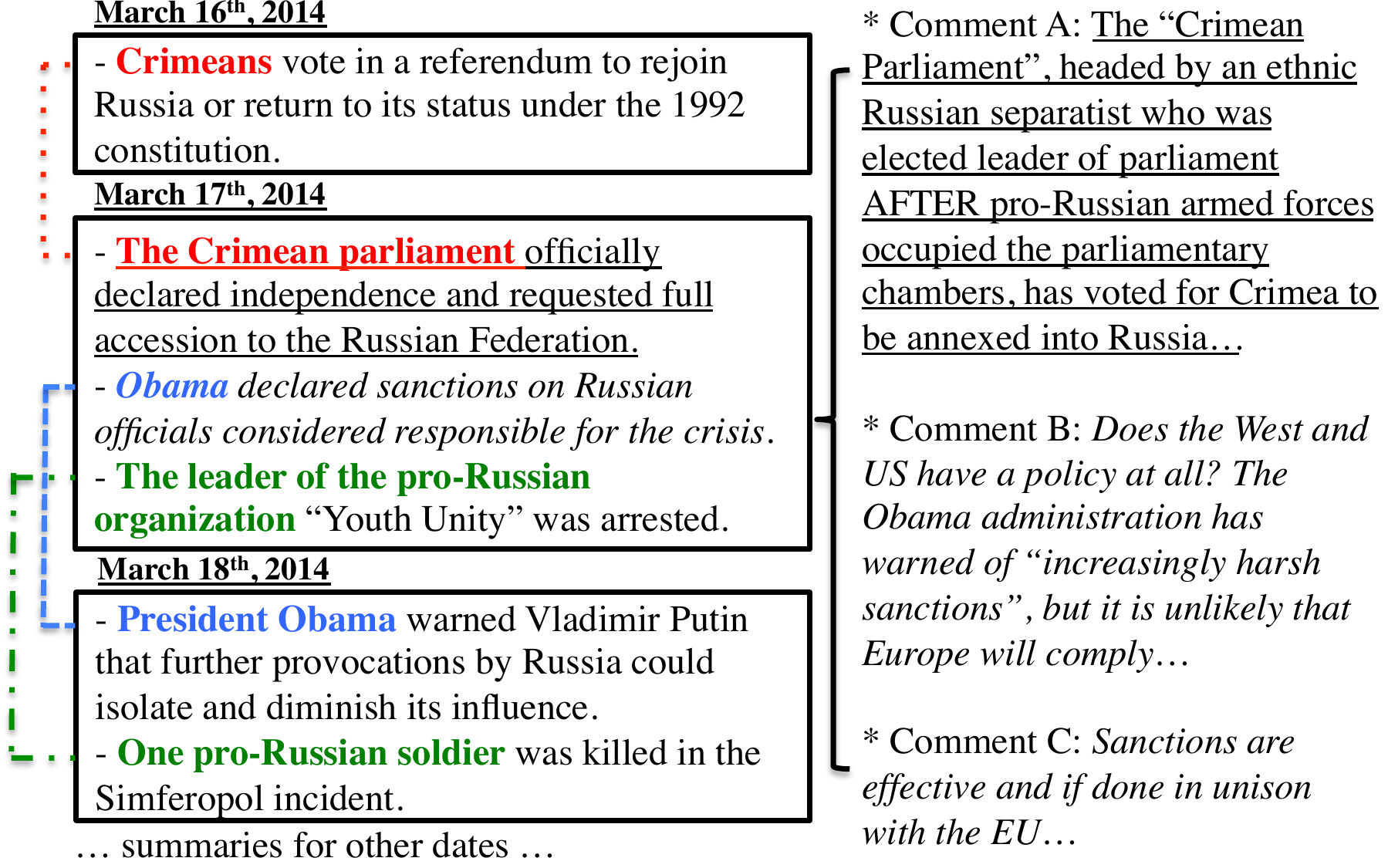}
\vspace{-5mm}
\caption{\fontsize{10}{11}\selectfont A snippet of the event timeline on Ukraine Crisis is displayed on the left. 
On the right, we display a set of representative comments addressing the article summary of March 17$^{th}$. Comment A (\underline{underlined}) brings a perspective on ``Crimean parliament passes declaration of independence'' (the article sentence is also underlined on the left). Comments B and C focus on Obama's sanctions on Ukrainian and Russian officials. Sentences linked by edges belong to the same event thread, which is centered on the entities with the same color.
}
\label{fig:timeline_intro}
\end{figure}

Social media sites on the Internet 
provide increasingly more, and increasingly popular, means for people to voice their opinions on trending events. 
Traditional news media --- the New York Times and CNN, for example ---
now provide online mechanisms that allow and encourage readers to share reactions, opinions, and personal experiences relevant to a news story. 
For complex emerging events, in particular, user comments can provide relevant, interesting and insightful information beyond the facts reported in the news. 
But their large volume and tremendous variation in quality make it impossible for readers to efficiently digest the user-generated content, much less integrate it with reported facts from the dozens or hundreds of news reports produced on the event each day.
In this work, we present a {\it socially-informed timeline generation system} that jointly generates a news article summary and a user comment summary for each day of an ongoing complex event. 
A sample (gold standard) timeline snippet for Ukraine Crisis is shown in Figure~\ref{fig:timeline_intro}. The event timeline is on the left; the comment summary for March {\small $17^{th}$} is on the right. 

While generating timelines from news articles and summarizing user comments have been studied as separate problems~\cite{Yan:2011:ETS:2009916.2010016,Ma:2012:TRC:2396761.2396798}, their joint summarization for timeline generation raises new challenges. 
Firstly, there should be a tight connection between the article and comment portion of the timeline. By definition, users comment on socially relevant events. So the important part of articles and insightful comments should both cover these events. 
Moreover, good reading experience requires that the article summary and comment summary demonstrate evident connectivity. For example, Comment C in Figure~\ref{fig:timeline_intro} (``Sanctions are effective and if done in unison with the EU") is obscure without knowing the context that ``sanctions are imposed by U.S''. Simply combining the outputs from a timeline generation system and a comment summarization system may lead to timelines that lack cohesion. 
On the other hand, articles and comments are from intrinsically different genres of text: articles emphasize facts and are written in a professional style; comments reflect opinions in a less formal way. Thus, it could be difficult to recognize the connections between articles and comments. 
Finally, it is also challenging to enforce continuity in timelines with many entities and events.

To address the challenges mentioned above, we formulate the timeline generation task as an optimization problem, where we maximize topic cohesion between the article and comment summaries while preserving their ability to reflect {\it important} concepts and subevents, adequate {\it coverage} of mentioned topics, and {\it continuity} of the timeline as it is updated with new material each day. We design a novel alternating optimizing algorithm that allows the generation of a high quality article summary and comment summary via mutual reinforcement. We demonstrate the effectiveness of our algorithm on four disparate complex event datasets collected over months from the New York Times, CNN, and BBC. Automatic evaluation using ROUGE~\cite{Lin:2003:AES:1073445.1073465} and gold standard timelines indicates that our system can effectively leverage user comments to outperform state-of-the-art approaches on timeline generation. In a human evaluation via Amazon Mechanical Turk, the comment summaries generated by our method were selected as the best in terms of informativeness and insightfulness in 66.7\% and 51.7\% of the evaluations (vs.\ 26.7\% and 30.0\% for randomly selected editor's-picks).


Especially, our optimization framework relies on two scoring functions that estimate the importance of including individual article sentences and user comments in the timeline. Based on the observation that entities or events frequently discussed in the user comments can help with identify summary-worthy content, we show that the scoring functions can be learned jointly by utilizing graph-based regularization. 
Experiments show that our joint learning model outperforms state-of-the-art ranking algorithms and other joint learning based methods when evaluated on sentence ranking and comment ranking. For example, we achieve an NDCG@3 of 0.88 on the Ukraine crisis dataset, compared to 0.77 from~\newcite{Yang:2011:SCS:2009916.2009954} which also conducts joint learning between articles and social context using factor graphs.

Finally, to encourage continuity in the generated timeline, we propose an entity-centered event threading algorithm. Human evaluation demonstrates that users who read timelines with event threads write more informative answers than users who do not see the threads while answering the same questions. This implies that our system constructed threads can help users better navigate the timelines and collect relevant information in a short time.


For the rest of the paper, we first describe data collection (Section~\ref{sec:data}). We then introduce the joint learning model for importance prediction (Section~\ref{sec:learning}). The full timeline generation system is presented in Section~\ref{sec:timeline}, which is followed by evaluations (Section~\ref{sec:result}). Related work and conclusion are in Sections~\ref{sec:related} and~\ref{sec:conclusion}.

\section{Data Collection and Preprocessing}
\label{sec:data}

We crawled news articles from New York Times (NYT), CNN, and BBC on four trending events: the missing Malaysia Airlines Flight MH370 (MH370), the political unrest in Ukraine (Ukraine), the Israel-Gaza conflict (Israel-Gaza), and the NSA surveillance leaks (NSA). For each event, we select a set of key words (usually entities' name), which are used to filter out irrelevant articles. We collect comments for NYT articles through NYT community API, and comments for CNN articles via Disqus API. \footnote{BBC comment volume is low, so we do not collect it.} 
NYT comments come with information on whether a comment is an editor's-pick. The statistics on the four datasets are displayed in Table~\ref{tab:statistics}.\footnote{The datasets are available at~\url{http://www.cs.cornell.edu/~luwang/data.html}.}

\begin{table}[ht]
\centering
    {\fontsize{9}{10}\selectfont
    \setlength{\baselineskip}{0pt}
    \setlength{\tabcolsep}{1.0mm}
    \begin{tabular}{|l|c|c|c|}
    \hline
	&\textbf{Time Span} & \textbf{\# Articles} & \textbf{\# Comments}\\ \hline
MH370 & 03/08 - 06/30 & 955 & 406,646\\ \hline
Ukraine & 03/08 - 06/30 & 3,779 & 646,961\\ \hline
Israel-Gaza& 07/20 - 09/30 & 909 & 322,244\\ \hline
NSA & 03/23 - 06/30 & 145 & 60,481\\ \hline

	\end{tabular}
	}
	\vspace{-3mm}
    \caption{\fontsize{10}{11}\selectfont Statistics on the four event datasets.}
	\label{tab:statistics}
\end{table}


We extract parse trees, dependency trees, and coreference resolution results of articles and comments with Stanford CoreNLP~\cite{manning-etal:2014:ACLDemo}. 
Sentences in articles are labeled with timestamps using SUTime~\cite{chang12}.

We also collect all articles with comments from NYT in 2013 (henceforth NYT2013) to form a training set for learning importance scoring functions on articles sentences and comments (see Section~\ref{sec:learning}). NYT2013 contains $3,863$ articles and $833,032$ comments.

\section{Joint Learning for Importance Scoring}
\label{sec:learning}
We first introduce a joint learning method that uses \textit{graph-based regularization} to simultaneously learn two functions --- a {\sc sentence} scorer and a {\sc comment} scorer --- that predict the importance of including an individual news article sentence or a particular user comment in the timeline. 

We train the model on the aforementioned NYT2013 dataset, where 20\% of the articles and their comments are reserved for parameter tuning. Formally, the training data consists of a set of articles {\small $D=\{d_{i}\}_{i=0}^{|D|-1}$}. Each article $d_{i}$ contains a set of sentences {\small $x_{s_{d_{i}}}=\{x_{s_{d_{i}}, j}\}_{j=0}^{|s_{d_{i}}|-1}$} and a set of associated comments {\small $x_{c_{d_{i}}}=\{x_{c_{d_{i}}, k}\}_{k=0}^{|c_{d_{i}}|-1}$}, where $|s_{d_{i}}|$ and $|c_{d_{i}}|$ are the numbers of sentences and comments for $d_{i}$. For simplicity, we use $x_{s}$ or $x_{c}$ to denote a sentence or a comment wherever there is no ambiguity. 

In addition, each article has a human-written abstract. We use the ROUGE-2~\cite{Lin:2003:AES:1073445.1073465} score of each sentence computed against the associated abstract as its gold-standard importance score. Each comment is assigned a gold-standard value of $1.0$ if it is an editor's pick, or $0.0$ otherwise.


The {\sc sentence} and {\sc comment} scorers rely on two classifiers, each designed to handle the special characteristics of news and user comments, respectively; and a graph-based regularizing constraint that encourages similarity between selected sentences and comments. We describe each component below.

\paragraph{Article {\sc sentence} Importance.}
Each sentence $x_{s}$ in a news article is represented as a $k$-dimensional feature vector {\small $\mathbf{x_{s}} \in \mathbb{R}^{k}$}, with a gold-standard label {\small $y_{s}$}. We denote the training set as a feature matrix {\small $\mathbf{\tilde{X}_{s}}$}, with a label vector {\small $\mathbf{\tilde{Y}_{s}}$}. 
%
To produce the {\sc sentence} scoring function {\small $f_{s}(x_{s})=\mathbf{x_{s}}\cdot \mathbf{w_{s}}$}, we use ridge regression to learn a vector {\small $\mathbf{w_{s}}$} that minimizes {\small $||\mathbf{\tilde{X}_{s}} \mathbf{w_{s}}-\mathbf{\tilde{Y}_{s}}||_{2}^{2} + \beta_{s}\cdot ||\mathbf{w_{s}}||_{2}^{2}$}. 
Features used in the model are listed in Table~\ref{tab:features_sent}. We also impose the following {\it position-based regularizing constraint} to encode the fact that the first sentence in a news article usually conveys the most essential information: 
{\small
$\lambda_{s}\cdot \sum_{d_{i}}\sum_{x_{s_{d_{i}}, j}, j\neq 0} ||(\mathbf{x_{s_{d_{i}}, 0}}-\mathbf{x_{s_{d_{i}}, j}})\cdot \mathbf{w_{s}}- (y_{s_{d_{i}}, 0}-y_{s_{d_{i}}, j})||_{2}^{2}$
}
, where {\small $x_{s_{d_{i}}, j}$} is the $j$-th sentence in document $d_{i}$. Term {\small $(\mathbf{x_{s_{d_{i}}, 0}}-\mathbf{x_{s_{d_{i}}, j}})\cdot \mathbf{w_{s}}$} measures the difference in predicted scores between the first sentence and any other sentence. This value is expected be close to the true difference. We further construct {\small $\mathbf{\tilde{X}^{\prime}_{s}}$} to contain all difference vectors {\small $(\mathbf{x_{s_{d_{i}}, 0}}-\mathbf{x_{s_{d_{i}}, j}})$}, with {\small $\mathbf{\tilde{Y}^{\prime}_{s}}$} as label difference vector. The objective function to minimize becomes

\vspace{-5mm}
{\small
\begin{equation}
\begin{split}
&J_{s}(\mathbf{w_{s}})=\\
&||\mathbf{\tilde{X}_{s}} \mathbf{w_{s}}-\mathbf{\tilde{Y}_{s}}||_{2}^{2} +\lambda_{s}\cdot ||\mathbf{\tilde{X}^{\prime}_{s}} \mathbf{w_{s}}-\mathbf{\tilde{Y}^{\prime}_{s}}||_{2}^{2}+ \beta_{s}\cdot ||\mathbf{w_{s}}||_{2}^{2}\\
\end{split}
\end{equation}
}

\vspace{-5mm}
\paragraph{User {\sc comment} Importance.}
Similarly, each comment $x_{c}$ is represented as an $l-$dimensional feature vector {\small $\mathbf{x_{c}}\in \mathbb{R}^{l}$}, with label {\small $y_{c}$}. Comments in the training data are denoted with a feature matrix {\small $\mathbf{\tilde{X}_{c}}$} with a label vector {\small $\mathbf{\tilde{Y}_{c}}$}. Likewise, we learn {\small $f_{c}(x_{c})=\mathbf{x_{c}}\cdot \mathbf{w_{c}}$} by minimizing 
{\small $||\mathbf{\tilde{X}_{c}} \mathbf{w_{c}} - \mathbf{\tilde{Y}_{c}}||_{2}^{2} + \beta_{c} \cdot ||\mathbf{w_{c}}||_{2}^{2}$}. 
%
%
%
%
%
%
Features are listed in Table~\ref{tab:features_comment}.  
We apply a {\it pairwise preference-based regularizing constraint} \cite{Joachims:2002:OSE:775047.775067} to incorporate a bias toward editor's picks: 
{\small $\lambda_{c}\cdot \sum_{d_{i}}\sum_{x_{c_{d_{i}}, j} \in \mathbf{E_{d_{i}}}, x_{c_{d_{i}}, k} \notin \mathbf{E_{d_{i}}}} ||(\mathbf{x_{c_{d_{i}}, j}}-\mathbf{x_{c_{d_{i}}, k}})\cdot \mathbf{w_{c}}- 1||_{2}^{2}$}
%
%
, where {\small $\mathbf{E_{d_{i}}}$} are the editor's picks for $d_{i}$. Term {\small $(\mathbf{x_{c_{d_{i}}, j}}-\mathbf{x_{c_{d_{i}}, k}})\cdot \mathbf{w_{c}}$} enforces the separation of editor's picks from regular comments. 
We further construct {\small$\mathbf{\tilde{X}^{\prime}_{c}}$} to contain all the pairwise differences {\small$(\mathbf{x_{c_{d_{i}}, j}}-\mathbf{x_{c_{d_{i}}, k}})$}. {\small$\mathbf{\tilde{Y}^{\prime}_{c}}$} is a vector of same size as {\small$\mathbf{\tilde{X}^{\prime}_{c}}$} with each element as $1$. Thus, the objective function to minimize is:

\vspace{-5mm}
{\small
\begin{equation}
\begin{split}
&J_{c}(\mathbf{w_{c}})=\\
&||\mathbf{\tilde{X}_{c}} \mathbf{w_{c}} - \mathbf{\tilde{Y}_{c}}||_{2}^{2} + \lambda_{c}\cdot ||\mathbf{\tilde{X}^{\prime}_{c}} \mathbf{w_{c}} - \mathbf{\tilde{Y}^{\prime}_{c}}||_{2}^{2}+\beta_{c}\cdot ||\mathbf{w_{c}}||_{2}^{2}
\end{split}
\end{equation}
}


\vspace{-5mm}
\paragraph{Graph-Based Regularization.}

The regularizing constraint is based on two mutually reinforcing hypotheses: (1) the importance of a sentence  depends partially on the availability of sufficient insightful comments that touch on topics in the sentence; (2) the importance of a comment depends partially on whether it addresses notable events reported in the sentences. 
For example, we want our model to bias {\small $\mathbf{w_{s}}$} to predict a high score for a sentence with high similarity to numerous insightful comments. 

We first create a bipartite graph from sentences and comments on the same articles, where edge \emph{weights} are based on the content similarity between a sentence and a comment
(TF-IDF similarity is used). Let {\small $\mathbf{\tilde{R}}$} be an {\small$N\times M$} adjacency matrix, where {\small $N$} and {\small $M$} are the numbers of sentences and comments. {\small$R_{sc}$} is the similarity between sentence $x_{s}$ and comment $x_{c}$. We normalize {\small $\mathbf{\tilde{R}}$} by {\small $\mathbf{\tilde{Q}}=\mathbf{\tilde{D}}^{-\frac{1}{2}}\mathbf{\tilde{R}}\mathbf{\tilde{D^{\prime}}}^{-\frac{1}{2}}$}, where {\small$\mathbf{\tilde{D}}$} and {\small$\mathbf{\tilde{D}^{\prime}}$} are diagonal matrices: {\small $\mathbf{\tilde{D}}\in\mathbb{R}^{N\times N}$, $D_{i,i}=\sum_{j=1}^{M}R_{i,j}$; $\mathbf{\tilde{D}^{\prime}}\in\mathbb{R}^{M\times M}$, $D^{\prime}_{j,j}=\sum_{i=1}^{N}R_{i,j}$.}
The interplay between the two types of data is encoded in the following regularizing constraint:

\vspace{-5mm}
{\small
\begin{equation}
\begin{split}
&J_{s, c}(\mathbf{w_{s}}, \mathbf{w_{c}})= \\
&\lambda_{sc}\cdot \sum_{d_{i}} \sum_{x_{s}\in x_{s_{d_{i}}}, x_{c}\in x_{c_{d_{i}}}} Q_{x_{s}, x_{c}}\cdot (\mathbf{x_{s}}\cdot\mathbf{w_{s}}-\mathbf{x_{c}}\cdot\mathbf{w_{c}})^{2}
\end{split}
\end{equation}
}




\vspace{-5mm}
\paragraph{Full Objective Function.}
Thus, the full objective function consists of the three parts discussed above:

\vspace{-5mm}
{\small
\begin{equation}
\label{eq:full}
\begin{split}
J(\mathbf{w_{s}}, \mathbf{w_{c}})=J_{s}(\mathbf{w_{s}})+J_{c}(\mathbf{w_{c}})+J_{s, c}(\mathbf{w_{s}}, \mathbf{w_{c}})
\end{split}
\end{equation}
}
Furthermore, using the following notation,
{\small
\[
\arraycolsep=1.4pt\def\arraystretch{1.0} 
\mathbf{\tilde{X}}=
\begin{bmatrix}
\mathbf{\tilde{X}_{s}} & \mathbf{0}\\
\mathbf{0} & \mathbf{\tilde{X}_{c}}\\
\end{bmatrix}
\,
\mathbf{\tilde{Y}}=
\begin{bmatrix}
\mathbf{\tilde{Y}_{s}}\\
\mathbf{\tilde{Y}_{c}}\\
\end{bmatrix}
\, 
\mathbf{\tilde{X}^{\prime}}=
\begin{bmatrix}
\mathbf{\tilde{X}^{\prime}_{s}} & \mathbf{0}\\
\mathbf{0} & \mathbf{\tilde{X}^{\prime}_{c}}\\
\end{bmatrix}
\,
\mathbf{\tilde{Y}^{\prime}}=
\begin{bmatrix}
\mathbf{\tilde{Y}^{\prime}_{s}}\\
\mathbf{\tilde{Y}^{\prime}_{c}}\\
\end{bmatrix}
\]
}
{\small
\[
\arraycolsep=1.4pt\def\arraystretch{1.0}
\hspace{-8mm} 
\tilde{\boldsymbol \beta}=
\begin{bmatrix}
\beta_{s} \mathbf{I_{k}} & \mathbf{0}\\
\mathbf{0} & \beta_{c} \mathbf{I_{l}}\\
\end{bmatrix}
\,
\hspace{2mm} 
\tilde{\boldsymbol \lambda}=
\begin{bmatrix}
\lambda_{s} \mathbf{I_{|X_{s}^{\prime}|}} & \mathbf{0}\\
\mathbf{0} & \lambda_{c} \mathbf{I_{|X_{c}^{\prime}|}}\\
\end{bmatrix}
\]
}
{\small
\[
\arraycolsep=1.4pt\def\arraystretch{1.0}
\hspace{-8mm} 
\mathbf{\tilde{L}}=
\begin{bmatrix}
\lambda_{sc} \mathbf{I_{|X_{s}|}} & -\lambda_{sc} \mathbf{\tilde{Q}}\\
-\lambda_{sc} \mathbf{\tilde{Q}^{T}} & \lambda_{sc} \mathbf{I_{|X_{c}|}}\\
\end{bmatrix}
\,
\hspace{2mm} 
\mathbf{w}=
\begin{bmatrix}
\mathbf{w_{s}}\\
\mathbf{w_{c}}\\
\end{bmatrix}
\]
}

\noindent
we can show a \textbf{closed form solution} to Equation~\ref{eq:full} as follows:

\vspace{-5mm}
{\small
\begin{equation}
\begin{split}
&\hat{\mathbf{w}}=\\
&(\mathbf{\tilde{X}^{T}\tilde{L}\tilde{X}+\tilde{X}^{T}\tilde{X}+\tilde{X}^{\prime T} \tilde{\boldsymbol \lambda} \tilde{X}^{\prime}}+\tilde{\boldsymbol \beta})^{-1}(\mathbf{\tilde{X}^{T}\tilde{Y}+\tilde{X}^{\prime T} \tilde{\boldsymbol \lambda} \tilde{Y}^{\prime}})
\end{split}
\end{equation}
}




\vspace{-5mm}
\begin{table}[ht]
\centering
    {\fontsize{9}{10}\selectfont
    \setlength{\baselineskip}{0pt}
    \setlength{\tabcolsep}{0.8mm}
    \begin{tabular}{|l|l|}
    \hline
	\underline{\textbf{Basic Features}}& \underline{\textbf{Social Features}}\\ 
	- num of words &  - avg/sum frequency of  \\
	- absolute/relative position &  words appearing in comment\\ 
	- overlaps with headline & - avg/sum frequency of  \\ 
	- avg/sum TF-IDF scores & dependency relations\\ 
	- num of NEs &   appearing in comment\\
	\hline
	\end{tabular}
	}
	\vspace{-3mm}
    \caption{\fontsize{10}{11}\selectfont Features used for sentence importance scoring.} 
	\label{tab:features_sent}
\end{table}

\begin{table}[ht]
    {\fontsize{9}{10}\selectfont
    \setlength{\baselineskip}{0pt}
    \setlength{\tabcolsep}{0.8mm}
    
	\hspace{-1mm}    
    \begin{tabular}{|l|l|}
    \hline
	\underline{\textbf{Basic Features}} &	\underline{\textbf{Readability Features}}\\ 
	- num of words & 	- Flesch-Kincaid Readability\\ 
	- num of sentences & 	- Gunning-Fog Readability\\
	- avg num of words   & 	\underline{\textbf{Discourse Features}}\\
	 ~~~per sentence &	-  num/proportion of connectives\\
	- num of NEs & 	- num/proportion of hedge words\\
	- num/proportion of   & 	\underline{\textbf{Article Features}}\\ 
	 ~~~capitalized words & 	- TF/TF-IDF simi with article\\
	- avg/sum TF-IDF &	- TF/TF-IDF simi with comments \\
	- contains URL & 	- JS/KL divergence (div) with article~\\
	- user rating (pos/neg)~ & 	- JS/KL div with comments\\

	\hline
	\end{tabular}
	
	\hspace{-1mm}    
    \begin{tabular}{|l|}
    \hline
    	\underline{\textbf{Sentiment Features}}\\
	- num /proportion of positive/negative/neutral words (MPQA\\
	 \cite{Wilson:2005:RCP}, General Inquirer~\cite{stone66})\\
	- num /proportion of sentiment words\\
	\hline
	\end{tabular}
	}
	\vspace{-3mm}
    \caption{\fontsize{10}{11}\selectfont Features used for comment importance scoring.} 
	\label{tab:features_comment}
\end{table}

\section{Timeline Generation}
\label{sec:timeline}
Now we present an optimization framework for timeline generation. 
Formally, for each day, our system takes as input a set of sentences {\small $V_{s}$} and a set of comments {\small $V_{c}$} to be summarized, and the (automatically generated) timeline {\small $\mathcal{T}$} (represented as threads) for days prior to the current day. It then identifies a subset {\small $S\subseteq V_{s}$} as the article summary and a subset {\small $C\subseteq V_{c}$} as the comment summary by maximizing the following function:

\vspace{-3mm}
{\small
\begin{equation}
\label{eq:full_timeline}
\mathcal{Z}(S, C; \mathcal{T})=\mathcal{S}_{qual}(S; \mathcal{T})+\mathcal{C}_{qual}(C)+\delta \mathcal{X}(S, C)
\end{equation}
}

\vspace{-3mm}
\noindent where {\small $\mathcal{S}_{qual}(S; \mathcal{T})$} measures the quality of the article summary {\small$S$} in the context of the historical timeline represented as event threads {\small $\mathcal{T}$}; 
{\small $\mathcal{C}_{qual}(C)$} computes the quality of the comment summary {\small$C$}; and 
{\small $\mathcal{X}(S, C)$} estimates the connectivity between {\small $S$} and {\small $C$}. 

We solve this maximization problem using an alternating optimization algorithm which is outlined in Section~\ref{sec:algorithm}. In general, we alternately search for a better article summary {\small $S$} with hill climbing search and a better comment summary {\small $C$} with Ford-Fulkerson algorithm until convergence.

In the rest of this section, we first describe an \textit{entity-centered event threading} algorithm to construct event threads $\mathcal{T}$ which are used to boost article timeline continuity. Then we explain how to compute {\small $\mathcal{S}_{qual}(S; \mathcal{T})$} and {\small $\mathcal{C}_{qual}(C)$} in Section~\ref{sec:quality}, followed by {\small $\mathcal{X}(S,C)$} in Section~\ref{sec:connection}.

\subsection{Entity-Centered Event Threading}
\label{sec:threading}
We present an event threading process where each thread connects sequential events centered on a set of relevant \emph{entities}. For instance, the following thread connects events about \emph{Obama}'s action towards the annexation of Crimea by \emph{Russia}:\\
{\fontsize{9}{10}\selectfont
\vspace{-1mm}
Day $1$: \emph{Obama} declared sanctions on \emph{Russian officials}.\\
\vspace{-1mm}
Day $2$: \emph{President Obama} warned \emph{Russian}.\\
\vspace{-1mm}
Day $3$: \emph{Obama} urges \emph{Russian} to move back its troops.\\
\vspace{-1mm}
Day $4$: \emph{Obama} condemns \emph{Russian} aggression in Ukraine.\\
}

We first collect relation extractions as \textit{(entity, relation, entity)} triples from OLLIE~\cite{Mausam:2012:OLL:2390948.2391009}, a dependency relation based open information extraction system. We retain extractions with confidence scores higher than $0.5$. 
We further design syntactic patterns based on~\newcite{Fader:2011:IRO:2145432.2145596} to identify relations expressed as a combination of a verb and nouns. Each relation contains at least one event-related word~\cite{Ritter:2012:ODE:2339530.2339704}.

The \textit{entity-centered event threading} algorithm works as follows: on the first day, each sentence in the summary becomes an individual cluster; thereafter, each sentence in the current day's article summary either gets attached to an existing thread or starts a new thread. The updated threads then become the input to next day's summary generation process. 
On day $n$, we have a set of threads {\small $\mathcal{T}=\{\tau: \mathbf{s_{1}},\mathbf{s_{2}},\cdots,\mathbf{s_{n-1}}\}$} constructed from previous $n-1$ days, where {\small $\mathbf{s_{i}}$} represents the set of sentences attached to thread $\tau$ from day $i$. The \textit{cohesion} between a new sentence {\small $s\in S$} and a thread $\tau$ is denoted as $cohn(s, \tau)$. $s$ is attached to $\hat{\tau}$ if there exists {\small$\hat{\tau}=\max_{\tau\in T} cohn(s, \tau)$} and {\small$cohn(s, \hat{\tau})>0.0$}.
Otherwise, $s$ becomes a new thread. 
We define {\small $cohn(s, \tau)=\min_{\mathbf{s_{i}}\in \tau, \mathbf{s_{i}}\neq \emptyset} tfsimi (\mathbf{s_{i}}, s)$}, where {\small $tfsimi(\mathbf{s_{i}}, s)$} measures the TF similarity between $\mathbf{s_{i}}$ and $s$.  We consider unigrams/bigrams/trigrams generated from the entities of our event extractions. 

\subsection{Summary Quality Measurement}
\label{sec:quality}
Recall that we learned two separate importance scoring functions for sentences and comments, which will be denoted here as {\small$imp_{s}(s)$} and {\small $imp_{c}(c)$}. 
With an article summary {\small$S$} and threads {\small$\mathcal{T}=\{\tau_{i}\}$}, the \textbf{article summary quality} function {\small$\mathcal{S}_{qual}(S;\mathcal{T})$} has the following form: 

{\footnotesize
$\mathcal{S}_{qual}(S;\mathcal{T})=\sum_{s\in S}imp(s)\\
+\theta_{cov}\sum_{s^{\prime}\in V_{s}} \min(\sum_{{s\in S}} tfidf(s, s^{\prime}), \alpha \sum_{\hat{s}\in V_{s}}tfidf(\hat{s}, s^{\prime}))\\
+\theta_{cont}\sum_{\tau \in \mathcal{T}} \max _{s_{k}\in S} cohn(s_{k}, \tau)$
}

{\small $tfidf(\cdot, \cdot)$} is the TF-IDF similarity function. {\small $\mathcal{S}_{qual}(S;\mathcal{T})$} captures three desired qualities of an article summary: \textit{importance} (first item), \textit{coverage} (second item), and the \textit{continuity} of the current summary to previously generated summaries. 
The coverage function has been used to encourage summary diversity and reduce redundancy~\cite{Lin:2011:CSF:2002472.2002537,wang-EtAl:2014:Coling2}. 
The continuity function considers how well article summary {\small$S$} can be attached to each event thread, thus favors summaries that can be connected to multiple threads. 

Parameters {\small $\theta_{cov}$} and {\small $\alpha$} are tuned on multi-document summarization dataset DUC 2003~\cite{Over2003}. Experiments show that system performance peaks and is stable for {\small $\theta_{cont}\in [1.0, 5.0]$}. We thus fix {\small $\theta_{cont}$} to $1.0$. We discard sentences with more than 80\% of content words covered by historical summaries. 
We use \textsc{Basic} to denote a system that only optimizes on importance and coverage (i.e. first two items in {\small $\mathcal{S}_{qual}(S;\mathcal{T})$}). The system optimizing {\small $\mathcal{S}_{qual}(S;\mathcal{T})$} is henceforth called \textsc{Thread}.

The \textbf{comment summary quality} function simply takes the form {\small$\mathcal{C}_{qual}(C)=\sum_{c\in C}imp_{c}(c)$}.

\subsection{Connectivity Measurement}
\label{sec:connection}
We encode two objectives in the connectivity function {\small$\mathcal{X}(S,C)$}: (1) encouraging topical cohesion (i.e. connectivity) between article summary and comment summary; and (2) favoring comments that cover diversified events.

\begin{figure}[th]
\hspace{-2mm}
\includegraphics[width=85mm,height=43mm]{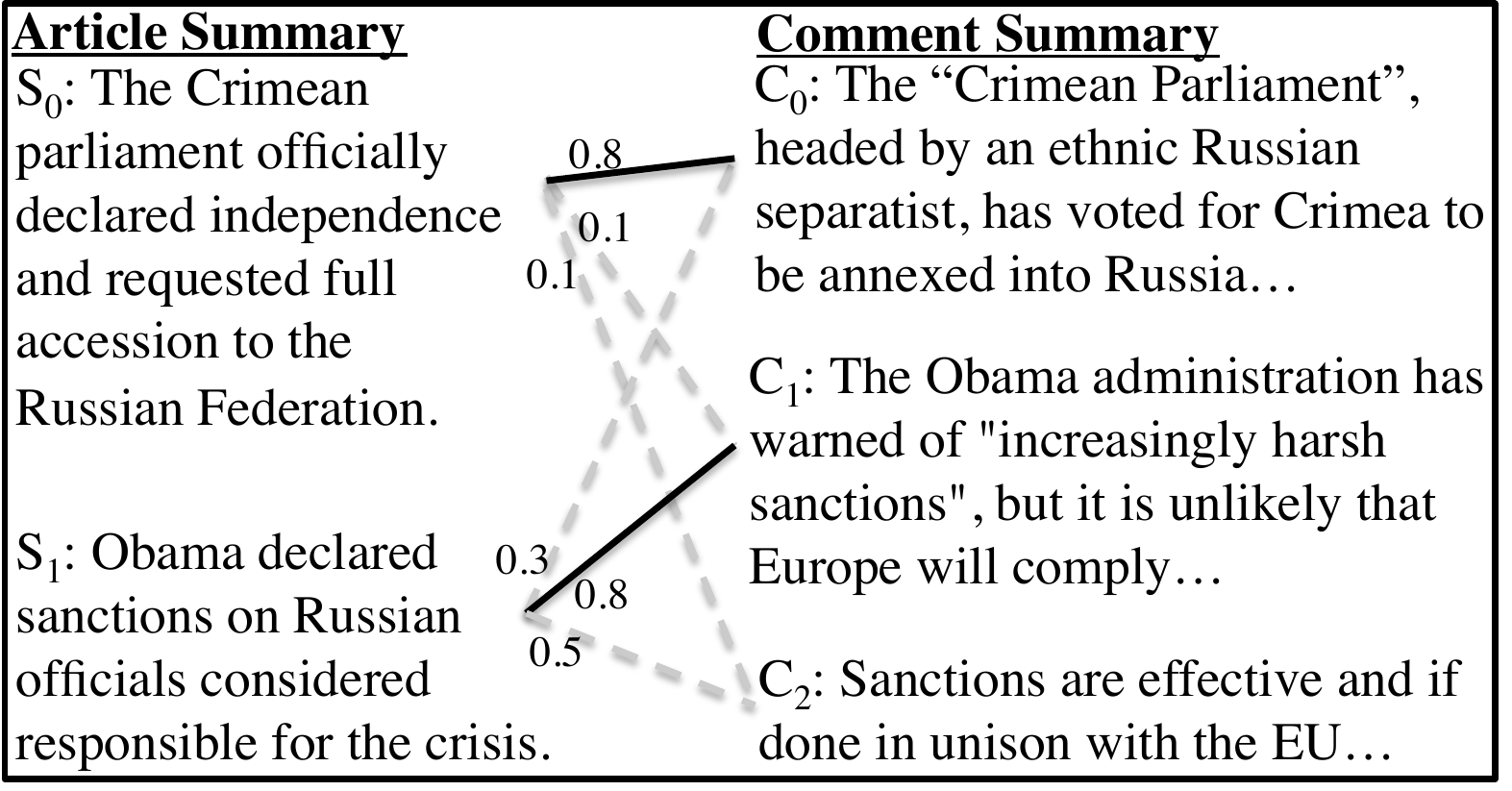}
\vspace{-7mm}
\caption{\fontsize{10}{11}\selectfont An example on computing the connectivity between an article summary (left) and a comment summary (right) via best matching in bipartite graph. Number on each edge indicates the content similarity between a sentence and a comment. Solid lines are edges in the best matching graph. For this example, the connectivity $\mathcal{X}(S, C)$ is $0.8+0.8=1.6$.}
\label{fig:bipartite_inference}
\end{figure}

Let {\small$conn(s, c)$} measure content similarity between a sentence {\small$s\in S$} and a comment {\small$c\in C$}. Connectivity between article summary {\small$S$} and comment summary {\small$C$} is computed as follows. We build a bipartite graph {\small$\mathcal{G}$} between {\small$S$} and {\small$C$} with edge weight as {\small$conn(s, c)$}. We then find an edge set {\small $\mathcal{M}$}, the best matching of {\small$\mathcal{G}$}. {\small$\mathcal{X}(S, C)$} is defined as the sum over edge weights in {\small $\mathcal{M}$}, i.e. {\small$\mathcal{X}(S, C)=\sum_{e\in \mathcal{M}} weight(e)$}. An example is illustrated in Figure~\ref{fig:bipartite_inference}.

We consider two options for $conn(s, c)$. One is \textit{lexical similarity} which is based on TF-IDF vectors. Another is \textit{semantic similarity}. Let {\small$R_{s}=\{(a_{s}, r_{s}, b_{s})\}$} and {\small$R_{c}=\{(a_{c}, r_{c}, b_{c})\}$} be the sets of dependency relations in $s$ and $c$. $conn(s, c)$ is calculated as:\\
\noindent {\scriptsize $\sum_{(a_{s}, r_{s}, b_{s})\in R_{s}}\max_{\substack{(a_{c}, r_{c}, b_{c})\in R_{c}\\ r_{s}=r_{c}}} simi(a_{s}, a_{c})\times simi(b_{s}, b_{c})$}

\noindent where {\small$simi(\cdot, \cdot)$} is a word similarity function. We experiment with shortest path based similarity defined on WordNet~\cite{Miller:1995:WLD:219717.219748} and Cosine similarity with word vectors trained on Google news~\cite{journals/corr/abs-1301-3781}. 
Systems using the three metrics that optimize {\small $\mathcal{Z}(S,C;\mathcal{T})$} are henceforth called \textsc{Thread+Opt$_{\tt TFIDF}$}, \textsc{Thread+Opt$_{\tt WordNet}$} and \textsc{Thread+Opt$_{\tt WordVec}$}.

\subsection{An Alternating Optimization Algorithm}
\label{sec:algorithm}
To maximize the full objective function {\small $\mathcal{Z}(S,C;\mathcal{T})$}, we design a novel alternating optimization algorithm (Alg.~\ref{alg:localsearch}) where we alternately find better {\small$S$} and {\small$C$}.

We initialize {\small$S_{0}$} by a greedy algorithm~\cite{Lin:2011:CSF:2002472.2002537} with respect to {\small$\mathcal{S}_{qual}(S;\mathcal{T})$}. Notice that {\small$\mathcal{S}_{qual}(S;\mathcal{T})$} is a submodular function, so that the greedy solution is a {\small $1-1/e$} approximation to the optimal solution of {\small $\mathcal{S}_{qual}(S;\mathcal{T})$}. 
Fixing {\small$S_{0}$}, we model the problem of finding {\small$C_{0}$} that maximizes {\small$\mathcal{C}_{qual}(C)+\delta \mathcal{X}(S_{0}, C)$} as a maximum-weight bipartite graph matching problem. This problem can be reduced to a maximum network flow problem, and then be solved by Ford-Fulkerson algorithm (details are discussed in~\cite{Kleinberg:2005:AD:1051910}). 
Thereafter, for each iteration, we alternately find a better {\small$S_{t}$} with regard to {\small$\mathcal{S}_{qual}(S;\mathcal{T})+\delta \mathcal{X}(S, C_{t-1})$} using hill climbing, and an exact solution {\small$C_{t}$} to {\small$\mathcal{C}_{qual}(C)+\delta \mathcal{X}(S_{t}, C)$} with Ford-Fulkerson algorithm. Iteration stops when the increase of {\small$\mathcal{Z}(S,C)$} is below threshold {\small$\epsilon$} (set to {\small $0.01$}). System performance is stable when we vary {\small$\delta\in [1.0, 10.0]$}, so we set {\small$\delta=1.0$}.

\begin{algorithm}
\small
\setstretch{0.6}
\SetKwInOut{Input}{Input}
\SetKwInOut{Output}{Output}
\hspace*{-3mm}  \Input{sentences $V_{s}$, comments $V_{c}$, threads $\mathcal{T}$, $\delta$, \\ \hspace*{-3mm} threshold $\epsilon$, functions $\mathcal{Z}(S,C;\mathcal{T})$, \\ \hspace*{-3mm} $\mathcal{S}_{qual}(S;\mathcal{T})$, $\mathcal{C}_{qual}(C)$, $\mathcal{X}(S,C)$}
\hspace*{-3mm}  \Output{article summary $S$, comment summary $C$}
\BlankLine

\tcc{\fontsize{8}{8}\selectfont Initialize $S$ and $C$ by greedy algorithm and Ford-Fulkerson algorithm}
\hspace*{-3mm} $S_{0}\leftarrow$  $\max_{S}\mathcal{S}_{qual}(S;\mathcal{T})$\;
\hspace*{-3mm} $C_{0}\leftarrow\max_{C} \mathcal{C}_{qual}(C)+\delta \mathcal{X}(S_{0}, C)$\;

\hspace*{-3mm} $t\leftarrow 1$\;
\hspace*{-3mm} $\Delta \mathcal{Z} \leftarrow \infty$\;
\hspace*{-3mm} \While{$\Delta \mathcal{Z} > \epsilon$}{
\hspace*{-3mm} \tcc{\fontsize{8}{8}\selectfont Step 1: Hill climbing algorithm}
\hspace*{-3mm} $S_{t}\leftarrow \max_{S} \mathcal{S}_{qual}(S;\mathcal{T})+\delta \mathcal{X}(S, C_{t-1})$\;
\hspace*{-3mm} \tcc{\fontsize{8}{8}\selectfont Step 2: Ford-Fulkerson algorithm}
\hspace*{-3mm} $C_{t}\leftarrow \max_{C}\mathcal{C}_{qual}(C)+\delta \mathcal{X}(S_{t}, C)$\;
\hspace*{-3mm} $\Delta \mathcal{Z}=\mathcal{Z}(S_{t}, C_{t};\mathcal{T})-\mathcal{Z}(S_{t-1}, C_{t-1};\mathcal{T})$\;
\hspace*{-3mm} $t\leftarrow t+1$\;
}
\caption{\fontsize{10}{11}\selectfont Generate article summary and comment summary for a given day via alternating optimization .}
\label{alg:localsearch}
\end{algorithm}

Algorithm~\ref{alg:localsearch} is guaranteed to find a solution at least as good as {\small$S_{0}$} and {\small$C_{0}$}. It progresses only if Step 1 finds {\small$S_{t}$} that improves upon {\small$\mathcal{Z}(S_{t-1},C_{t-1};\mathcal{T})$}, and Step 2 finds {\small$C_{t}$} where {\small$\mathcal{Z}(S_{t},C_{t};\mathcal{T})\geq \mathcal{Z}(S_{t},C_{t-1};\mathcal{T})$}. 

\section{Experimental Results}
\label{sec:result}
\subsection{Evaluation of {\sc sentence} and {\sc comment} Importance Scorers}
We test importance scorers (Section~\ref{sec:learning}) on single document \textit{sentence ranking} and \textit{comment ranking}. 

For both tasks, we compare with two previous systems on joint ranking and summarization of news articles and tweets. \textit{\newcite{Yang:2011:SCS:2009916.2009954}} employ supervised learning based on factor graphs to model content similarity between the two types of data. We use the same features for this model. \textit{\newcite{Gao:2012:JTM:2396761.2398417}} summarize by including the complementary information between articles and tweets, which is estimated by an unsupervised topic model.\footnote{We thank Zi Yang and Peng Li for providing the code.} 
We also consider two state-of-the-art rankers: \textit{RankBoost}~\cite{Freund:2003:EBA:945365.964285} and \textit{LambdaMART}~\cite{citeulike:9923905}. Finally, we use a \textit{position baseline} that ranks sentences based on their position in the article, and a \textit{rating baseline} that ranks comments based on positive user ratings.

\begin{figure}[ht]
\subfloat
{
	\hspace{-6mm}
    \includegraphics[width=48mm,height=35mm]{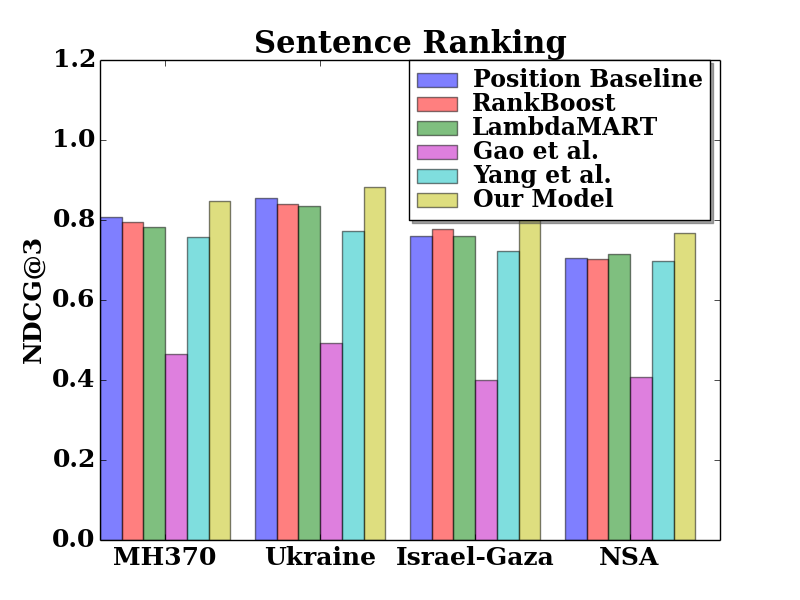}
}
\subfloat
{	\hspace{-7mm}
    \includegraphics[width=48mm,height=35mm]{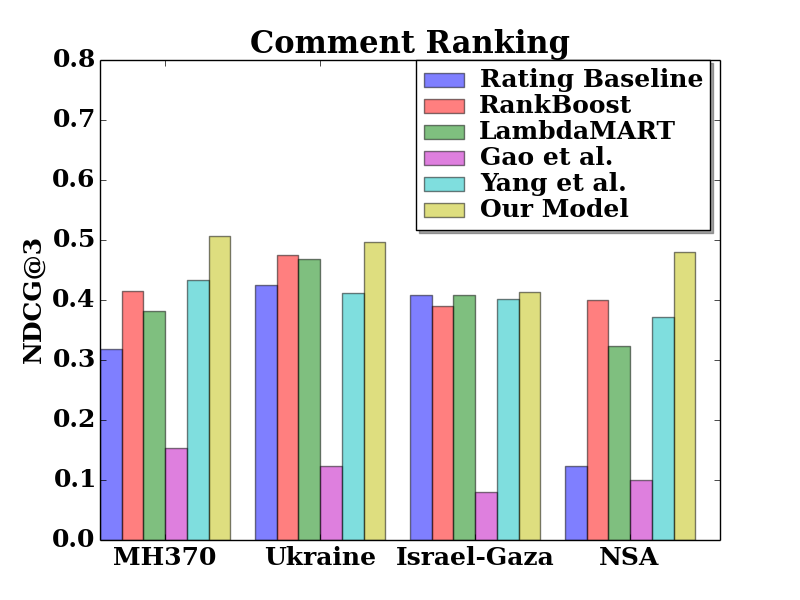}
}
\vspace{-3mm}
\caption{\fontsize{10}{11}\selectfont Evaluation of sentence and comment ranking on the four datasets by using normalized discounted cumulative gain at top 3 returned results (NDCG@3). Our joint learning based approach uniformly outperforms all the other comparisons.}
\label{fig:result_Ranking}
\end{figure}

We evaluate using normalized discounted cumulative gain at top 3 returned results (NDCG@3). Sentences are considered relevant if they have ROUGE-2 scores larger than $0.0$ (computed against human abstracts), and comments are considered relevant if they are editor's picks.\footnote{We experiment with all articles for sentence ranking, and NYT comments (with editor's picks) for comment ranking.} 
Figure~\ref{fig:result_Ranking} demonstrates that our joint learning model uniformly outperforms all the other comparisons for both ranking tasks. In general, supervised learning based approaches (e.g. our method, \newcite{Yang:2011:SCS:2009916.2009954}, RankBoost, and LambdaMART) produce better results than unsupervised method (e.g. \newcite{Gao:2012:JTM:2396761.2398417}).\footnote{Similar results are obtained with mean reciprocal rank.}

\vspace{-5mm}

\subsection{Leveraging User Comments}
\label{sec:autoEval}
In this section, we test if our system can leverage comments to produce better article-based summaries for event timelines. We collect \textbf{gold-standard timelines} for each of the four events from the corresponding Wikipedia page(s), NYT topic page, or BBC news page.

We consider two existing timeline creation systems that only utilize news articles, and a timeline generated from single-article human abstracts: (1) \textsc{\newcite{Chieu:2004:QBE:1008992.1009065}} select sentences with high ``interestingness'' and ``burstiness" using a likelihood ratio test to compare word distributions of sentences with articles in neighboring days. 
(2) \textsc{\newcite{Yan:2011:ETS:2009916.2010016}} design an evolutionary summarization system that selects sentences based on on coverage, coherence, and diversity. 
(3) We construct a timeline from the human \textsc{Abstract}s provided with each article: we sort them chronologically according to article timestamps and add abstract sentences into each daily summary until reaching the word limit.

We test on five variations of our system. The first two systems generate article summaries with no comment information by optimizing {\footnotesize $S_{qual}(S; \mathcal{T})$} using a greedy algorithm: \textsc{Basic} ignores event threading; \textsc{Thread} considers the threads. 
\textsc{Thread+Opt$_{\tt TFIDF}$}, \textsc{Thread+Opt$_{\tt WordNet}$} and \textsc{Thread+Opt$_{\tt WordVec}$} (see Section~\ref{sec:connection}) leverage user comments to generate article summaries as well as comment summaries based on alternating optimization of Equation 3. 
Although comment summaries are generated, they are not used in the evaluation.

For all systems, we generate daily article summaries of at most 100 words, and select 5 comments for the corresponding comment summary. We employ ROUGE~\cite{Lin:2003:AES:1073445.1073465} to automatically evaluate the content coverage (in terms of ngrams) of the article-based timelines vs. gold-standard timelines. ROUGE-2 (measures bigram overlap) and ROUGE-SU4 (measures unigram and skip-bigrams separated by up to four words) scores are reported in Table~\ref{tab:timeline_rouge}. As can be seen, under the alternating optimization framework, our systems, employing both articles and comments, consistently yield better ROUGE scores than the three baseline systems and our systems that do not leverage comments. Though constructed from single-article abstracts, baseline \textsc{Abstract} is found to contain redundant information and thus limited in content coverage. This is due to the fact that different media tend to report on the same important events.

\begin{table}[ht]
    {\fontsize{8}{9}\selectfont
    \setlength{\baselineskip}{0pt}
    \setlength{\tabcolsep}{0.6mm}
    \hspace{-2mm}	
    \begin{tabular}{|l|cc|cc|cc|cc|}
    \hline
	&\multicolumn{2}{|c|}{\textbf{MH370}}&\multicolumn{2}{|c|}{\textbf{Ukraine}}&\multicolumn{2}{|c|}{\textbf{Israel-Gaza}}&\multicolumn{2}{|c|}{\textbf{NSA}}\\

    \textit{}&\textit{\fontsize{7}{7}\selectfont R-2}&\textit{\fontsize{7}{7}\selectfont R-SU4}&\textit{\fontsize{7}{7}\selectfont R-2}&\textit{\fontsize{7}{7}\selectfont R-SU4}&\textit{\fontsize{7}{7}\selectfont R-2}&\textit{\fontsize{7}{7}\selectfont R-SU4}&\textit{\fontsize{7}{7}\selectfont R-2}&\textit{\fontsize{7}{7}\selectfont R-SU4}\\ \hline

\textsc{Chieu and Lee} & 6.43 & 10.89 & 4.64 & 8.87 & 3.38& \textbf{7.32} & 6.14 & 9.73\\ 
\textsc{Yan et al.} & 6.37 & 10.35 & 4.57 & 8.67 & 2.39& 5.78 & 3.99 & 7.73\\ 
\textsc{Abstract}& 6.16 & 10.62 & 3.85& 8.40& 2.21& 5.42 & 7.03 & 8.65\\ \hline \hline

\multicolumn{9}{|l|}{- \textit{Greedy Algorithm}}\\
\textsc{Basic}& 6.59 & 9.80 & 5.31& 9.23& 3.15& 6.20& 3.81& 7.58\\ 
\textsc{Thread}& 6.55& 10.86& 5.73 & 9.75& 3.16& 6.16& 6.29& 10.09\\ 
\multicolumn{9}{|l|}{- \textit{Alternating Optimization (leveraging comments)}}\\
\textsc{Thread+Opt$_{\tt TFIDF}$}& \textit{8.74} & 11.63& \textit{9.10}& \textit{12.59}& 3.78 & 6.45& \textit{8.07} & \textit{10.31}\\ 
\textsc{Thread+Opt$_{\tt WordNet}$} & \textit{8.73} & \textbf{\textit{11.87}}& \textit{8.67}& \textit{12.10}& \textbf{4.11} & 6.64& \textbf{\textit{8.63}}& \textbf{\textit{11.12}}\\ 
\textsc{Thread+Opt$_{\tt WordVec}$} & \textbf{\textit{9.29}} & 11.63 & \textbf{\textit{9.16}}& \textbf{\textit{12.72}}& 3.75& 6.38& \textit{8.29}& \textit{10.36}\\ 

	\hline
	\end{tabular}
	}
	\vspace{-3mm}
    \caption{\fontsize{10}{11}\selectfont ROUGE-2 (R-2) and ROUGE-SU4 (R-SU4) scores (multiplied by 100) for different timeline generation approaches on four event datasets. Systems that statistically significantly
outperform the three baselines ($p < 0.05$, paired $t-$test) are in \textit{italics}. Numbers in \textbf{bold} are the highest score for each column.} 
	\label{tab:timeline_rouge}
\end{table}

\subsection{Evaluating Socially-Informed Timelines}
We evaluate the full article+comment-based timelines on Amazon Mechanical Turk. Turkers are presented with a timeline consisting of five consecutive days' article summaries and four variations of the accompanying comment summary: {\sc random}ly selected comments, {\sc user's-picks} (ranked by positive user ratings), \emph{randomly} selected {\sc editor's-picks} and timelines produced by the \textsc{Thread+Opt$_{\tt WordVec}$} version of {\sc our system}. We also include one noisy comment summary (i.e.\ irrelevant to the question) to avoid spam. We display two comments per day for each system.\footnote{For our system, we select the two comments with highest importance scores from the comment summary.}

Turkers are asked to rank the comment summary variations according to \textit{informativeness} and \textit{insightfulness}. For informativeness, we ask the Turkers to judge based only on
knowledge displayed in the timeline, and to rate each comment summary based on how much relevant information they learn from it. For insightfulness, Turkers are required to focus on insights and valuable opinions. They are requested to leave a short explanation of their ranking.

15 five-day periods are randomly selected. We solicit four distinct Turkers located in the U.S. to evaluate each set of timelines. An inter-rater agreement of Krippendorff's $\alpha$ of 0.63 is achieved for informativeness ranking and $\alpha$ is 0.50 for insightfulness ranking.


\begin{table}[ht]
\centering
    {\fontsize{10}{11}\selectfont
     \setlength{\tabcolsep}{1mm}
	\begin{tabular}{|l|c|c|c|c|}
    \hline
	&\multicolumn{2}{|c|}{\textbf{Informativeness}}&\multicolumn{2}{|c|}{\textbf{Insightfulness}}\\
	& \% Best  & Avg Rank& \% Best & Avg Rank\\ \hline
	Random & 1.7\% & 3.67 & 3.3\% & 3.58 \\ \hline
	User's-picks & 5.0\% & 2.83 & 15.0\% & 2.55\\ \hline
	Editor's-picks & 26.7\% & 2.05 & 30.0\% & 2.22\\ \hline
	Our system & \textbf{66.7\%} & \textbf{1.45} & \textbf{51.7\%} & \textbf{1.65}\\ \hline
	
	\end{tabular}
	}
	\vspace{-3mm}
    \caption{\fontsize{10}{11}\selectfont Human evaluation results on the comment portion of socially-informed timelines. \textbf{Boldface} indicates statistical significance vs.\ other results in the same column using a Wilcoxon signed-rank test ($p<0.05$). On average, the output from our system is ranked higher than all other alternatives.}
	\label{tab:human_comment}
\end{table}

Table~\ref{tab:human_comment} shows the percentage of times a particular method is selected as producing the best comment portion of the timeline, as well as the micro-average rank of each method, for both informativeness and insightfulness. Our system is selected as the best in 66.7\% of the evaluations for informativeness and 51.7\% for insightfulness.  In both cases, we statistically significantly outperform ($p<0.05$ using a Wilcoxon signed-rank test) the editor's-picks and user's-picks. 
Turkers' explanations indicate that they prefer our comment summaries mainly because they are ``very informative and insightful to what was happening'', and ``show the sharpness of the commenter''. Turkers sometimes think the summaries randomly selected from editor's-picks ``lack connection'', and characterize user's-picks as ``the information was somewhat limited''.

Figure~\ref{fig:example_timeline} shows part of the timeline generated by our system for the Ukraine crisis.

\begin{figure}[ht]
	{\fontsize{8.5}{9}\selectfont
	\setlength{\tabcolsep}{0.6mm}
	\hspace{-2mm}
	\begin{tabular}{|p{40mm}|p{40mm}|}
	\hline
	\textbf{Article Summary} & \textbf{Comment Summary}\\ \hline
	\underline{2014-03-17}
	Obama administration froze the U.S. assets of seven Russian officials, while similar sanctions were imposed on four Ukrainian officials. $\ldots$
	& Theodore Roosevelt said that the worst possible thing you can do in diplomacy is ``soft hitting''. That is what the US and the EU are doing in these timid ``sanctions'' against people without any overseas accounts$\ldots$\\ \hline
	\underline{2014-03-18}  	
	Ukraine does not recognize a treaty signed in Moscow on Tuesday making its Crimean peninsula a part of Russia$\ldots$
	&Though there were many in Crimea who supported annexation, there were certainly some who did not. what about those people?$\ldots$\\ \hline
	\underline{2014-03-19}  
	The head of NATO warned on Wednesday that Russian President Vladimir Putin may not stop with the annexation of Crimea $\ldots$
	& If you look at a real map , Crimea is an island and has always been more connected to Russia than to Ukraine$\ldots$\\ \hline
	\underline{2014-03-20}  
	The United States on Thursday expanded its sanctions on Russians$\ldots$ 
	&The US and EU should follow up economic sanctions with concrete steps to strengthen NATO$\ldots$  \\ \hline
	\end{tabular}
	}
\vspace{-3mm}
\caption{\fontsize{10}{11}\selectfont A snippet of timeline generated by our system \textsc{Thread+Opt$_{\tt WordVec}$} for the Ukraine crisis. Due to space limitations, we only display partial summaries.}
\label{fig:example_timeline}
\end{figure}

\subsection{Human Evaluation of Event Threading}
Here we evaluate on the utility of event threads for high-level information access guidance: \textit{can event threads allow users to easily locate and absorb information with a specific interest in mind?}

We first sample a 10-day timeline for each dataset from those produced by the  \textsc{Thread+Opt$_{\tt WordVec}$} variation of our system. We designed one question for each timeline. Sample questions are: ``describe the activities for searching for the missing flight MH370", and ``describe the attitude and action of Russian Government on Eastern Ukraine''. 
We recruited 10 undergraduate and graduate students who are native speakers of English. Each student first read one question and its corresponding timeline for 5 minutes. The timeline was then removed, and the student wrote down an answer for the question. We asked each student to answer the question for each of four timelines (one for each event dataset). Two timelines are displayed with threads, and two without threads. We presented threads by adding a thread number in front of each sentence. 

We then used Amazon Mechanical Turk to evaluate the informativeness of students' answers. Turkers were asked to read all 10 answers for the same question, with five answers based on timelines with threads and five others based on timelines without threads. After that, they rated each answer with an informativeness score on a 1-to-5 rating scale (1 as ``not relevant to the query'', and 5 as ``very informative''). We also added two quality control questions. 
%
Table~\ref{tab:human_summary} shows that the average rating for answers written after reading timelines \textit{with threads} is {\small $3.29$} ({\small$43\%$} are rated {\small$\geq 4$}), higher than the {\small $2.58$} for the timelines with \textit{no thread} exhibited ({\small$30\%$} are rated {\small$\geq 4$}). 

\begin{table}[th]
\centering
    {\fontsize{10}{11}\selectfont
    \setlength{\tabcolsep}{1mm}
	\begin{tabular}{|l|c|c|c|c|}
	\hline
	\textbf{Answer Type}& \textbf{Avg $\pm$ STD} & \textbf{Rated 5 (\%)} & \textbf{Rated 4 (\%)}\\ \hline
No Thread & 2.58  $\pm$ 1.20 & 7\% & 23\%\\ \hline
With Threads & 3.29 $\pm$ 1.28 & 17\% & 26\%\\ \hline

	\end{tabular}
	}
	\vspace{-3mm}
    \caption{\fontsize{10}{11}\selectfont Human evaluation on the  informativeness of answers written after reading timelines \textit{with threads} vs. with \textit{no thread}. Answers written with access to threads are rated higher ($3.29$) than the ones with no thread ($2.58$).}
	\label{tab:human_summary}
\end{table}

\vspace{-4mm}
\section{Related Work}
\label{sec:related}
There is a growing interest in generating article summaries informed by social context. Existing work focuses on learning users' interests from comments 
and incorporates the learned information into a news article summarization system~\cite{Hu:2008:CDS:1390334.1390385}. 
\newcite{Zhao:2013:TGS:2484028.2484103} instead estimate word distributions from tweets, and bias a Page Rank algorithm to give higher restart probability to sentences with similar distributions. 
Generating tweet+article summaries has been recently investigated in~\newcite{Yang:2011:SCS:2009916.2009954}. They propose a factor graph to allow sentences and tweets to mutually reinforce each other. \newcite{Gao:2012:JTM:2396761.2398417} exploit a co-ranking model to identify sentence-tweet pairs with complementary information estimated from a topic model. These efforts handle a small number of documents and tweets, while we target a larger scale of data.



In terms of timeline summarization, the \newcite{Chieu:2004:QBE:1008992.1009065} system ranks sentences according to ``burstiness'' and ``interestingness" estimated by a likelihood ratio test. \newcite{Yan:2011:ETS:2009916.2010016} explore an optimization framework that maximizes the relevance, coverage, diversity, and coherence of the timeline. Neither system has leveraged the social context. 
%
%
Our event threading algorithm is also inspired by work on topic detection and tracking (TDT)~\cite{Allan98b}, where efforts are made for document-level link detection and topic tracking. 
Similarly, \newcite{Nallapati:2004:ETW:1031171.1031258} investigate event threading for articles, where they predict linkage based on causal and temporal dependencies. 
\newcite{Shahaf:2012:TTG:2187836.2187957} instead seek for connecting articles into one coherent graph. 
To the best of our knowledge, we are the first to study sentence-level event threading.

\section{Conclusion}
\label{sec:conclusion}
We presented a socially-informed timeline generation system, which constructs timelines consisting of article summaries and comment summaries. An alternating optimization algorithm is designed to maximize the connectivity between the two sets of summaries as well as their importance and information coverage. Automatic and human evaluations showed that our system produced more informative timelines than state-of-the-art systems. Our comment summaries were also rated as very insightful.

\section*{Acknowledgments}
We thank John Hessel, Lillian Lee, Moontae Lee, David Lutz, Karthik Raman, Vikram Rao, Yiye Ruan, Xanda Schofield, Adith Swaminathan, Chenhao Tan, Bishan Yang, other members of Cornell NLP group, and the NAACL reviewers for valuable suggestions and advice on various aspects of this work. This work was supported in part by DARPA DEFT Grant FA8750-13-2-0015.

\bibliographystyle{naaclhlt2015}

\end{document}